\documentclass[]{article}
\usepackage[letterpaper]{geometry}
\usepackage{mtsummit2021}
\usepackage{times}
\usepackage{url}
\usepackage{latexsym}
\usepackage{natbib}
\usepackage{layout}
\usepackage{adjustbox}
\usepackage{CJKutf8}
\usepackage{amssymb}
\usepackage{pifont}
\usepackage{comment}

\usepackage{todonotes}
\usepackage{colortbl}
\usepackage{graphicx,multirow}

\newcommand{\iiitt}{\texttt{IIITT}}
\newcommand{\onlenlp}{\texttt{oneNLP-IIITH}}
\newcommand{\adapt}{\texttt{adapt\_dcu}}
\newcommand{\ucf}{\texttt{UCF}}
\newcommand{\athree}{\texttt{A3108}}
\newcommand{\cfilt}{\texttt{CFILT-IITBombay}}

\begin{document}

\title{\bf Findings of the LoResMT 2021 Shared Task on COVID and Sign Language for Low-resource Languages}

\author{\name{\bf Atul Kr. Ojha\textsuperscript{1,2}} \hfill
\addr{atulkumar.ojha@insight-centre.org}\\
\name{\bf Chao-Hong Liu\textsuperscript{3}}\hfill
\addr{ch.liu@acm.org}\\
\name{\bf Katharina Kann\textsuperscript{4}} \hfill \addr{katharina.kann@colorado.edu}\\
\name{\bf John Ortega\textsuperscript{5}}\hfill
\addr{jortega@cs.nyu.edu}\\
\name{\bf Sheetal Shatam\textsuperscript{2}} \hfill
\addr{panlingua@outlook.com}\\
\name{\bf Theodorus Fransen\textsuperscript{1}} \hfill \addr{theodorus.fransen@insight-centre.org}\\

\addr{\textsuperscript{1}Data Science Institute, NUIG, Galway\\
\textsuperscript{2}Panlingua Language Processing LLP, New Delhi}\\
\textsuperscript{3}Potamu Research Ltd\\
\textsuperscript{4}University of Colorado at Boulder\\
\textsuperscript{5}New York University \\ 
}

\maketitle
\pagestyle{empty}

\begin{abstract}
We present the findings of the LoResMT 2021 shared task which focuses on machine translation (MT) of COVID-19 data for both low-resource spoken and sign languages. The organization of this task was conducted as part of the fourth workshop on technologies for machine translation of low resource languages (LoResMT). Parallel corpora is presented and publicly available which includes the following directions: English$\leftrightarrow$Irish, English$\leftrightarrow$Marathi, and Taiwanese Sign language$\leftrightarrow$Traditional Chinese. Training data consists of 8112, 20933 and 128608 segments, respectively. There are additional monolingual data sets for Marathi and English that consist of 21901 segments. The results presented here are based on entries from a total of eight teams. Three teams submitted systems for English$\leftrightarrow$Irish while five teams submitted systems for English$\leftrightarrow$Marathi. Unfortunately, there were no systems submissions for the Taiwanese Sign language$\leftrightarrow$Traditional Chinese task. Maximum system performance was computed using BLEU and follow as 36.0 for English--Irish, 34.6 for Irish--English, 24.2 for English--Marathi, and 31.3 for Marathi--English.
\end{abstract}

\section{Introduction}
The workshop on technologies for machine translation of low resource languages (LoResMT)\footnote{\url{https://sites.google.com/view/loresmt/}} is a yearly workshop which focuses on scientific research topics and technological resources for machine translation (MT) using low-resource languages.
Based on the success of its three predecessors \citep{ws-2018-amta-2018-technologies,karakanta2019proceedings,loresmt-2020-technologies}, the fourth LoResMT workshop intoduces a shared task section based on COVID-19 and sign language data as part of its research objectives. The hope is to provide assistance with translation for low-resource languages where it could be needed most during the COVID-19 pandemic.

To provide a trajectory of the LoResMT shared task success, a summary of the previous tasks follows. The first LoResMT shared task \citep{karakanta2019proceedings} took place in 2019. There, monolingual and parallel corpora for Bhojpuri, Magahi, Sindhi, and Latvian were provided as training data for two types of machine translation systems: neural and statistical. As an extension to the first shared task, a second shared task \citep{ojha-etal-2020-findings} was presented in 2020 which focused on zero-shot approaches for MT systems.

This year, the shared task introduces a new objective focused on MT systems for COVID-related texts and sign language. Participants for this shared task were asked to submit novel MT systems for the following language pairs:
 \begin{itemize}
     \item English$\leftrightarrow$Irish
     \item English$\leftrightarrow$Marathi
     \item Taiwanese Sign Language$\leftrightarrow$Traditional Chinese
 \end{itemize}
The low-resource languages presented in this shared task were found to be sufficient data for baseline systems to perform translation on the latest COVID-related texts and sign language. Irish, Marathi, and Taiwanese Sign Language can be considered low-resource languages and are translated to either English or traditional Chinese -- their high-resource counterpart. 

The rest of our work is organized as follows. Section \ref{sec:Task} presents the setup and schedule of the shared task. Section \ref{sec:Data} presents the data set used for the competition. Section \ref{sec:Participants} describes the approaches used by participants in the competition and Section \ref{sec:Results} presents and analyzes the results obtained by the competitors. Lastly, in Section \ref{sec:Conclusion} a conclusion is presented along with potential future work.

\section{Shared task setup and schedule}
\label{sec:Task}

This section describes how the shared task was organized along with the systems. Registered participants were sent links to the training, development, and/or monolingual data (refer to Section \ref{sec:Data} for more details). They were allowed to use additional data to train their system with the condition that any additional data used should be made publicly available. Participants were moreover allowed to use pre-trained word embeddings and linguistic models that are publicly available. 
As a manner of detecting which data sets were used during training, participants were given the following markers for denotation:
\begin{itemize}
    \item \emph{``-a"} - Only provided development, training and monolingual corpora.
    \item \emph{``-b"}- Any provided corpora, plus publicly available language's corpora and pre-trained/linguistic model (e.g. systems used pre-trained word2vec, UDPipe, etc. model).
   \item \emph{``-c"} - Any provided corpora, plus any publicly external monolingual corpora.
\end{itemize}

\noindent Each team was allowed to submit any number of systems for evaluation and their best 3 systems were included in the final ranking presented in this report. Each submitted system was evaluated on standard automatic MT evaluation metrics; 
BLEU~\citep{papineni2002bleu}, CHRF~\citep{popovic-2015-chrf} and TER~\citep{post-2018-call}.

The schedule for deliver of training data and release of test data along with notification and submission can be found in Table \ref{tab:timeline}.

\begin{table}[!ht]
\centering 
    \begin{tabular}{|l|l|}
\hline
   {\bf Date}  & {\bf  Event}  \\ \hline
 May 10, 2021 & Release of training data   \\
\hline
July 01, 2021   &  Release of test data  \\
\hline
July 13, 2021  & Submission of the systems   \\
\hline
July 20, 2021   &  Notification of results  \\
\hline
July 27, 2021   &  Submission of shared task papers  \\
\hline
August 01, 2021   &  Camera-ready  \\
\hline
    \end{tabular}

\caption{LoResMT 2021 Shared Task programming}
\label{tab:timeline}
\end{table}
\newpage

\section{Languages and data sets}
\label{sec:Data}
In this section, we present background information about the languages and data sets featured in the shared task along with a itemized view of the linguistic families and number of segments in Table~\ref{dataset}.

\subsection{Training data set}
\begin{itemize}
    \item \textbf{English$\leftrightarrow$Irish}
   Irish (also known as Gaeilge) has around 170,000 L1 speakers and ``1.85 million (37\%) people across the island (of Ireland) claim to be at least somewhat proficient with the language''. In the Republic of Ireland, it is the national and first official language. It is also one of the official languages of the European Union and a recognized minority language in Northern Ireland with the ISO \textit{ga} code.\footnote{\url{https://cloud.dfki.de/owncloud/index.php/s/sAs23JKXRwEEacn}}\\
   \\
    English-Irish bilingual COVID sentences/documents were extracted and aligned from the following sources: (a) Gov.ie\footnote{\url{www.gov.ie}} - Search for services or information , 
    (b) Ireland's Health Services\footnote{\url{https://www.hse.ie/}} - HSE.ie ,
    (c) Revenue Irish Taxes and Customs\footnote{\url{https://www.revenue.ie/}} and
        (d) Europe Union\footnote{\url{https://europa.eu}}. In addition, the Irish bilingual training data was built from monolingual data using back translation \citep{sennrich-etal-2016-improving}. English and Irish monolingual data was compiled from Wikipedia pages and newspapers such as The Irish Times\footnote{\url{https://www.irishtimes.com/}}, RTE\footnote{\url{https://www.rte.ie/news/} \& \url{https://www.rte.ie/gaeilge/}} and COVID-19 pandemic in the Republic of Ireland\footnote{\url{https://en.wikipedia.org/wiki/COVID-19_pandemic_in_the_Republic_of_Ireland}}. Back-translated and crawled data were cross-validated for accuracy by language experts leaving approximately 8,112 Irish parallel sentences for the training data set. 
    \item \textbf{English$\leftrightarrow$Marathi}
    Marathi, which has the ISO code \textit{mr}, is dominantly spoken in India's Maharashtra state. It has around 83,026,680 speakers.\footnote{\url{https://censusindia.gov.in/2011Census/C-16_25062018_NEW.pdf}} It belongs to the Indo-Aryan language family.\\
    \\
    English--Marathi parallel
    COVID sentences were extracted from the Government of India website and online newspapers such as PMIndia\footnote{\url{https://www.pmindia.gov.in/}}, myGOV\footnote{\url{https://www.mygov.in/}}, Lokasatta\footnote{\url{https://www.loksatta.com/}}, BBC Marathi and English\footnote{\url{https://www.bbc.com/marathi} \& \url{https://www.bbc.com/}}. After pre-processing and manual validation, approximately 20,993 parallel training sentences were left. Additionally, English and Marathi monolingual sentences were crawled from the online newspapers and Wikipedia (see Table \ref{dataset}). 
    
    \item \textbf{Taiwanese Sign Language $\leftrightarrow$ Traditional Chinese}
    According to UN, there are ``72 million deaf people worldwide... they use more than 300 different sign languages.''\footnote{\url{https://www.un.org/en/observances/sign-languages-day}} In Taiwan, Taiwanese Sign Language is a recognized national language, with a population of less than thirty thousand ``speakers''. Taiwanese Sign Language (and Korean Sign Language) evolved from Japanese Sign Language and share about 60\% of ``words'' between them.\\
    The sign language data set is prepared from press conferences for COVID-19 response, which were held daily or weekly depending on the pandemic situation in Taiwan. Fig. 1 shows a sample video of sign language and its translations in Traditional Chinese (excerpted from the corpus) and English.
    \end{itemize}
 \begin{figure}[hb]
    \includegraphics[width=.24\textwidth]{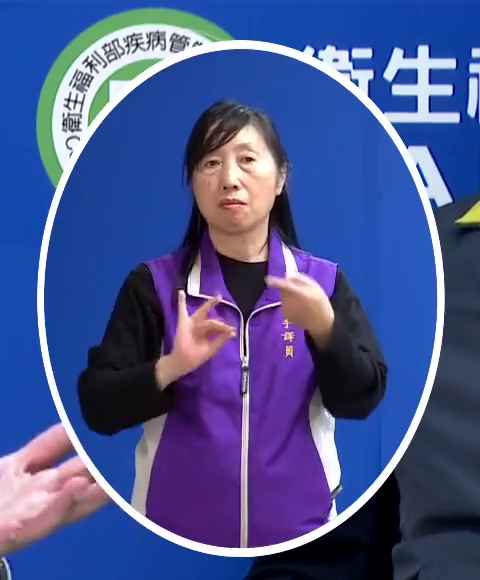}\hfill
    \includegraphics[width=.24\textwidth]{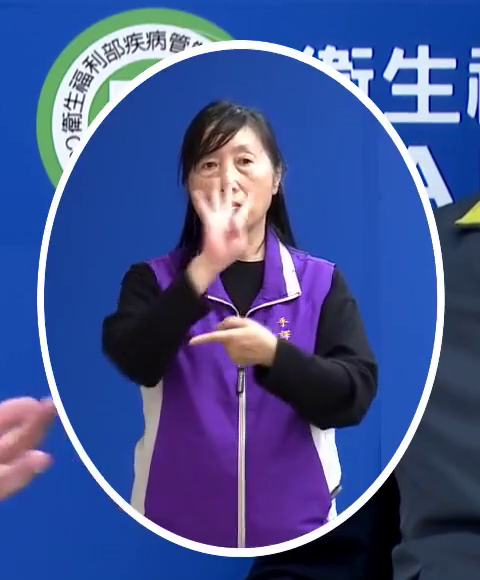}\hfill
    \includegraphics[width=.24\textwidth]{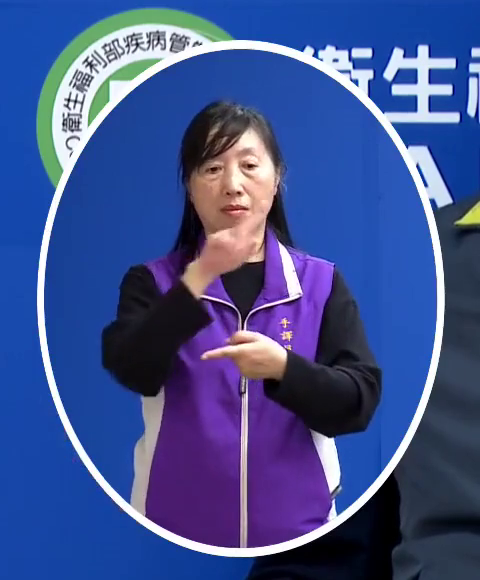}\hfill
    \includegraphics[width=.24\textwidth]{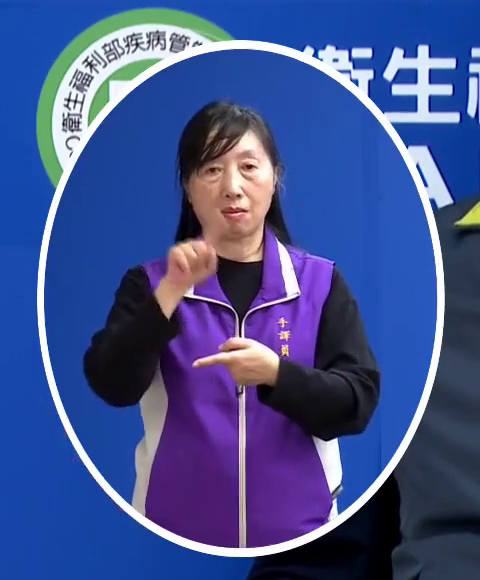}\hfill
    \includegraphics[width=.24\textwidth]{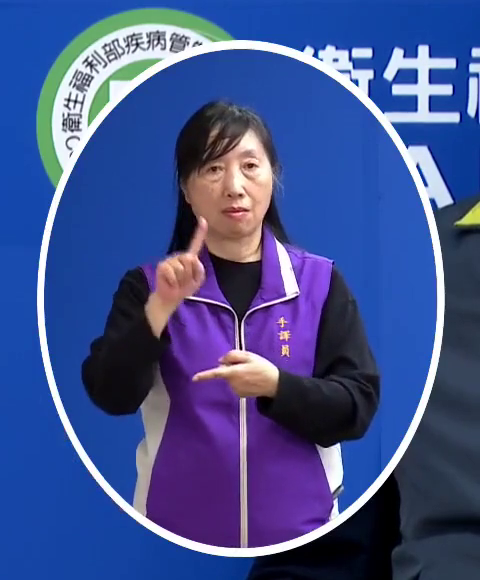}\hfill
    \includegraphics[width=.24\textwidth]{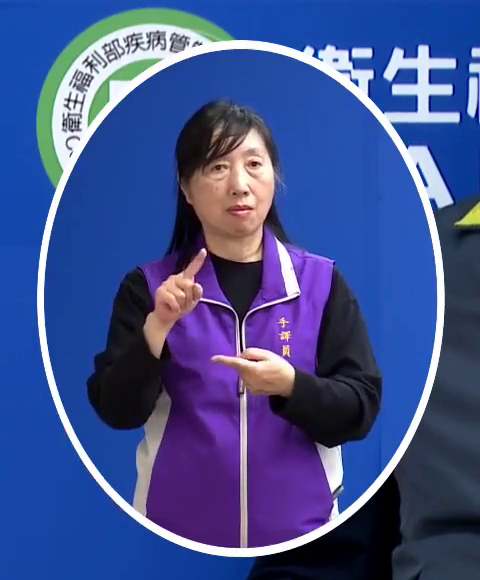}\hfill
    \includegraphics[width=.24\textwidth]{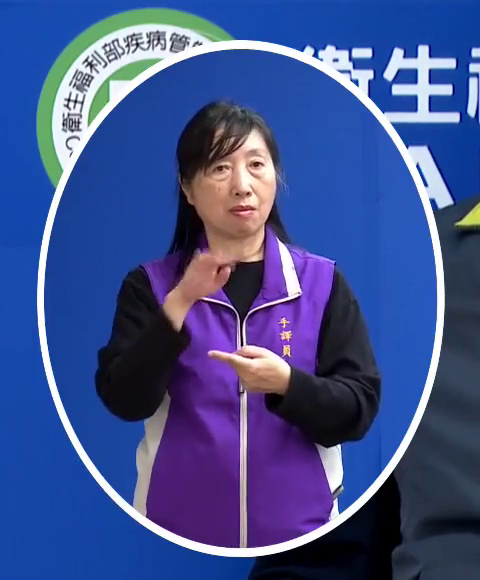}\hfill
    \includegraphics[width=.24\textwidth]{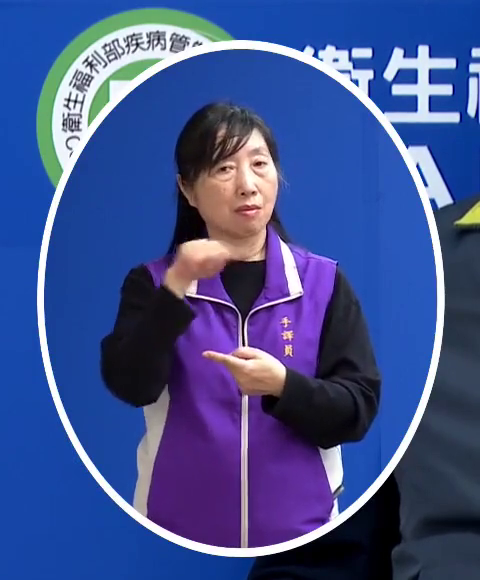}\hfill
    \includegraphics[width=.24\textwidth]{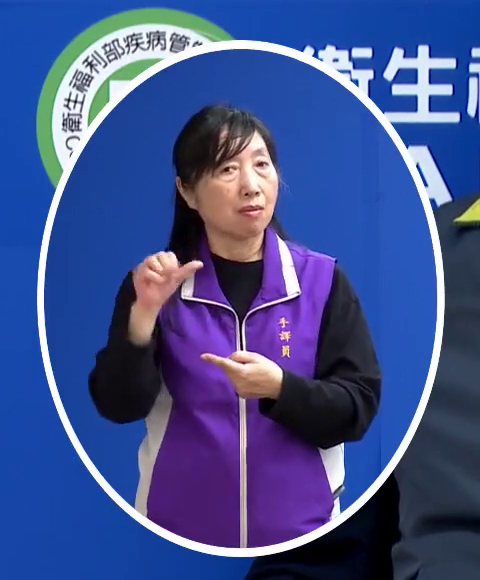}\hfill
    \includegraphics[width=.24\textwidth]{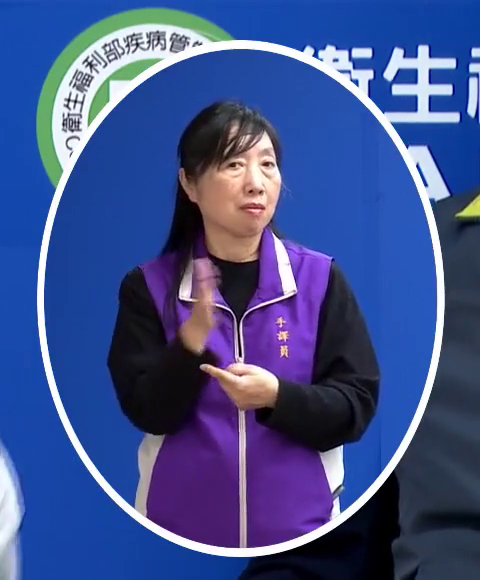}\hfill
    \includegraphics[width=.24\textwidth]{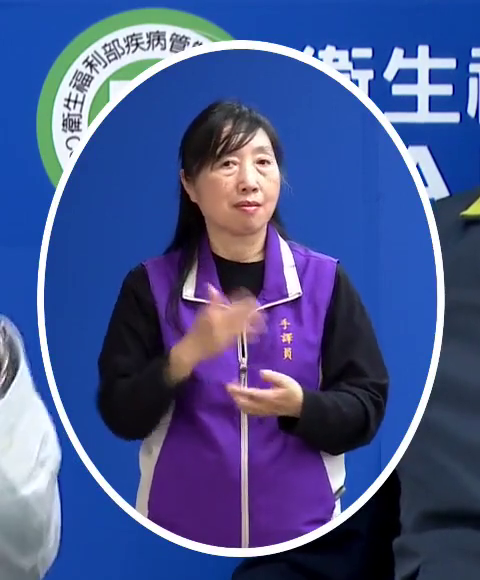}\hfill
    \includegraphics[width=.24\textwidth]{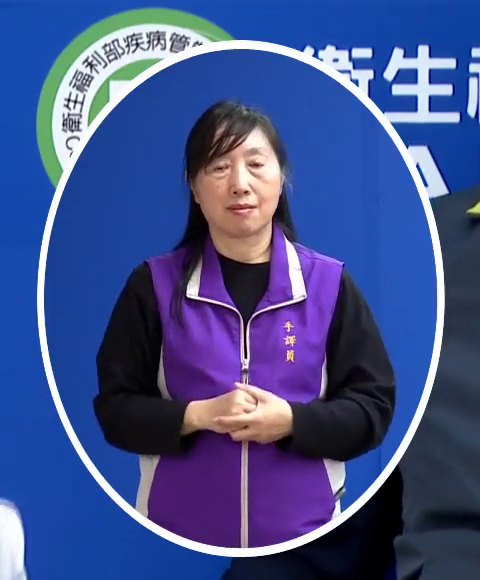}\hfill
    \caption{Sample of a sign language video in frames (excerpted from C00207\_711.mp4 in the corpus; Translations in Traditional Chinese: ``\begin{CJK*}{UTF8}{bsmi}4.1吧 或4.5  大概是這樣的一個比例\end{CJK*}'', and English: ``The ratio is approximately 4.1 or 4.5'')}
    \label{fig:signlg}
\end{figure}

\newpage
\subsection{Development and test data sets}
Similar to the training data, English-Irish and English-Marathi language pair’s dev and test data sets were crawled from bilingual and/or monolingual websites. Additionally, some parallel segments and terminology were taken from the Translation Initiative for COVID-19~\citep{anastasopoulos-etal-2020-tico}, a manually translated and validated data set created by professional translators and native speakers of the target languages. The participants of the shared task were provided with the manual translations of which 502 Irish and 500 Marathi development segments were used while 250 (Irish-English), 500 (English-Irish), 500 (English-Marathi) and 500 ( Marathi-English) manually translated segments were used for testing. Taiwanese Sign Language $\leftrightarrow$ Traditional Chinese language pair's participants were provided with 3071 segments and videos for development and 7,053 videos for sign language testing.

The detailed statistics of the data set in each language is provided in Table\ref{dataset}.  The complete shared task data sets are available publicly\footnote{\url{https://github.com/loresmt/loresmt-2021}}.

\begin{table}[!ht]
    \centering
    \small
    \setlength{\tabcolsep}{3.5pt}

    \begin{tabular}{|l|l|l|r|r|r|r|}
        \hline
        \textbf{Language} & \textbf{Code} & \textbf{Family} & \bf Train &\textbf{Dev} & \bf Monolingual & \textbf{Test} \\
        \hline
        English & en & Indo-Germanic & - & - & 8,826 & - \\
        \hline
        Irish & ga & Celtic & 8112 & 502 & - & 750\\
        \hline
        Marathi & mr & Indo-Aryan  & 20,933 & 500 & 21,902 & 1,000\\
        \hline
        TSign & sgTW & Japanese Sign Language & 128,608 & 3,071 & - & 7,053 \\
        \hline
        TChinese & zhTW & Mandarin Chinese & 128,608 & 3,071 & - & 7,053 \\
        \hline
    \end{tabular}
    \caption{Statistics of the Shared task data (TSign refers to Taiwanese Sign Language and TChinese refers to Traditional Chinese)}
    \label{dataset}
\end{table}

\section{Participants and methodology}
\label{sec:Participants}
A total of 12 teams registered for the shared task: 5 teams registered to participate for all language pairs, 5 teams registered to participate only for English$\leftrightarrow$Marathi, one team registered for Taiwanese$\leftrightarrow$Mandarin (Traditional Chinese) sign language and one team registered for English$\leftrightarrow$Irish. Out of these, a total of 6 teams submitted their systems on COVID while none of them submitted a system for sign language. Out of the submitted systems, two teams participated for the English$\leftrightarrow$Irish 
and English$\leftrightarrow$Marathi tasks, one team participated for English-Irish and three teams participated for English$\leftrightarrow$Marathi (see Table \ref{tab:teams}). All the teams who submitted their systems were invited to submit system description papers describing their experiments. Table~\ref{tab:teams} identifies the participating teams and their language choices.

\begin{table*}[ht!]
\centering
\adjustbox{max width=\textwidth}{
    \begin{tabular}{|c|c|c|c|c|}
   \hline
   {\bf Team}  & {\bf English--Irish} & {\bf English--Marathi} & {\bf TSign--TChinese} & {\bf System Description Paper}\\
   \hline
\iiitt & en2ga \& ga2en & en2mr \& mr2en & --- & \citep{puranik-etal-2021-attentive} \\
\hline
\onlenlp  & --- & en2mr \& mr2en & --- & \citep{mujadia-sharma-2021-english} \\
\hline
\athree   & --- & en2mr \& mr2en & --- & \citep{yadav-shrivastava-2021-a3}\\
\hline
\cfilt  & --- & en2mr \& mr2en & --- & \citep{jain-etal-2021-evaluating}\\
\hline
\ucf  & en2ga \& ga2en & en2mr \& mr2en & --- & \citep{chen-fazio-2021-the}\\ 
\hline
\adapt & en2ga & --- & --- & \citep{lankford-etal-2021-machine} \\ 
\hline
    {\bf Total} & {\bf 3 } & {\bf 5} & {\bf 0 } &{\bf 6 }\\
\hline
    \end{tabular}
}
\caption{Details of the teams and submitted systems for the LoResMT 2021 Shared Task.}
\label{tab:teams}
\end{table*}

Next, we give a short description of the approaches used by each team to build their systems. More details about the approaches can be found in the papers by respective teams in the accompanying proceeding.

\begin{itemize}
    \item {\bf IIITT}~\citep{puranik-etal-2021-attentive} used a fairseq pre-trained model Indictrans for English-Marathi. It consists of two models that can translate from Indic to English and vice-versa. The model can perform 11 languages: Assamese, Bengali, Gujarati, Hindi, Kannada, Malayalam, Marathi, Oriya, Punjabi, Tamil, Telugu pre-trained on the Samanantar data set, the largest data set for Indic languages during the time of submission. The model is fine-tuned on the training data set provided by the organizers and a parallel bible corpus for Marathi. The team used the parallel bible parallel corpus from a previous task (MultiIndicMT task in WAT 2020). After conducting various experiments, the best checkpoint was recorded and predicted upon. For Irish, the team fine-tuned an Opus MT model from Helsinki NLP on the training data set, and then predicted results after recording. After careful experimentation, the team observed that the Opus MT model outperformed the other models giving it the highest scoring model award.
    \item {\bf oneNLP-IIITH}~\citep{mujadia-sharma-2021-english} used a sequence to sequence neural model with a transformer network (4 to 8 layers) with label smoothing and dropouts to reduce overfitting with English-Marathi and Marathi-English. The team explored the use of different linguistic features like part-of-speech and morphology on sub-word units for both directions. In addition, the team explored forward and backward translation using web-crawled monolingual data.
    \item {\bf A3108}~\citep{yadav-shrivastava-2021-a3} built a statistical machine translation (smt) system in both directions for English$\leftrightarrow$Marathi language pair. Its initial baseline experiments used various tokenization schemes to train models. By using optimal tokenization schemes, the team was able to create synthetic data and train an augmented dat aset to create more statistical models. Also, the team reordered English syntax to match Marathi syntax and further trained another set of baseline and data augmented models using various tokenization schemes.
    \item {\bf CFILT-IITBombay}~\citep{jain-etal-2021-evaluating} buildt three different neural machine translation systems; a baseline English--Marathi system, a Baseline Marathi-English system, and a English--Marathi system that was based on back translation. The team explored the performance of the NMT systems between English and Marathi languages. Also, they explored the performance of back-translation using data obtained from NMT systems trained on a very small amount of data. From their experiments, the team observed that back-translation helped improve the MT quality over the baseline for English-Marathi.
    \item {\bf UCF}~\citep{chen-fazio-2021-the} used transfer learning, uni-gram and sub-word segmentation methods for English--Irish, Irish--English, English--Marathi and Marathi--English. The team conducted their experiment using an OpenNMT LSTM system. Efforts were constrained by using transfer learning and sub-word segmentation on small amounts of training data. Their models achieved the following BLEU scores when constraining on tracks of English--Irish, Irish--English, and Marathi--English: 13.5, 21.3, and 17.9, respectively.
    \item {\bf adapt\_dcu}~\citep{lankford-etal-2021-machine} used a transformer training approach carried out using OpenNMT-py and sub-word models for English--Irish. The team also explored domain adaptation techniques while using a Covid-adapted generic 55k corpus, fine-tuning, mixed fine-tuning and combined data set approaches were compared with models trained on an extended in-domain data set.
\end{itemize}
\section{Results}
\label{sec:Results}
As discussed, participants were allowed to use data sets other than those provided. The best three results for English-Irish, Irish-English, English-Marathi and Marathi-English language pairs are presented in Tables \ref{tab:results1} and \ref{tab:results2}. The complete submitted systems results are available publicly\footnote{\url{https://github.com/loresmt/loresmt-2021}}. Table \ref{tab:results1} depicts how the \ucf and team were able to gain the highest and lowest results for Irish-English and English-Marathi with shared data. The highest scores were 21.3 BLEU, 0.45 CHRF and 0.711 TER, while the lowest scores were 5.1 BLEU, 0.22 CHRF and 0.872 TER. However, with the additional data and by using pre-trained models (see Table \ref{tab:results2}), \adapt and achieved the best results for English-Irish where scores were 36 BLEU, 0.6 CHRF and 0.531 TER. Contrastingly, \ucf  and scored the lowest for English-Marathi. The lowest scores were 4.8 BLEU, 0.29 CHRF and 1.063 TER.

\begin{table*}[ht!]
\centering
    \begin{tabular}{|l|c|c|c|c|}
    \hline
  {\textbf{Team}} & \textbf{System/task description} & \textbf{BLEU} & \textbf{CHRF} & \textbf{TER}\\
  \hline
  \adapt & en2ga-a & 9.8 & 0.34 & 0.880 \\
  \hline
  \ucf & ga2en-TransferLearning-a & 21.3 & 0.45 & 0.711 \\
  \hline
 \cfilt & en2mr-Backtranslation-a &  12.2 & 0.38 & 0.979 \\
 \hline
 \cfilt & en2mr-Baseline\_200-a & 11 & 0.38 & 0.961 \\
  \hline
  \cfilt & en2mr-Baseline\_1600-a & 10.8 & 0.38 & 0.935 \\
  \hline
  \onlenlp & en2mr-Method1-a &  10.4 & 0.32 & 0.907\\
  \hline
  \athree & en2mr-Method29transliterate-a & 11.8  & 0.45 & 0.95\\
  \hline
  \athree & en2mr-Method29unk-a & 11.8  & 0.45 & 0.95\\
  \hline
  \athree & en2mr-Method10unk-a & 11.4  & 0.43 & 0.934\\
  \hline 
  \ucf & en2mr-UnigramSegmentation-a &  5.1 & 0.22 &  0.872\\
    \hline
    \cfilt  & mr2en-Baseline\_1000-a & 16.6 & 0.41 & 0.870\\
    \hline
    \cfilt  & mr2en-Baseline\_1200-a & 16.3 & 0.40 & 0.867\\
    \hline
    \cfilt  & mr2en-Baseline\_1400-a & 16.2 & 0.41 & 0.879\\
    \hline
    \onlenlp  &  mr2en-Method1-a & 16.7 & 0.40 & 0.835\\
  \hline
  \onlenlp  &  mr2en-Method2-a & 16.2 & 0.41 & 0.831\\
  \hline
 \athree & mr2en-Method7transliterate-a & 14.6  & 0.47 & 0.945\\
  \hline
  \athree & mr2en-Method7unk-a & 14.6  & 0.47 & 0.945\\
  \hline
  \athree & mr2en-Method20transliterate-a & 14.5  & 0.42 & 0.866\\
  \hline
  \ucf & mr2en-UnigramSegmentation-a & 17.9 & 0.40 & 0.744\\
  \hline
    \end{tabular}
\caption{Results of submitted systems at English$\leftrightarrow$Irish \& English$\leftrightarrow$Marathi in the ``-a'' method}
\label{tab:results1}
\end{table*}
\begin{table*}[!ht]
\centering
    \begin{tabular}{|l|c|c|c|c|}
    \hline
  {\textbf{Team}} & \textbf{System/task description} & \textbf{BLEU} & \textbf{CHRF} & \textbf{TER}\\
  \hline
  \adapt  & en2ga-b & 36.0 & 0.60 & 0.531\\
  \hline
   \iiitt  & en2ga-helsnikiopus-b & 25.8 & 0.53 & 0.629\\
  \hline
  \iiitt  & ga2en-helsinkiopus-b & 34.6 & 0.61 & 0.586\\
  \hline
  \iiitt & en2mr-IndicTrans-b & 24.2 & 0.59 & 0.597 \\
  \hline
  \onlenlp & en2mr-Method2-c & 22.2 & 0.56 & 0.746 \\
  \hline
 \onlenlp & en2mr-Method3-c & 22.0 & 0.56 & 0.753 \\
  \hline
 \onlenlp  & en2mr-Method1-c & 21.5  & 0.56 & 0.746\\
  \hline
 \ucf  & en2mr-UnigramSegmentation-b & 4.8  & 0.29 & 1.063\\
  \hline 
 \onlenlp  & mr2en-Method3-c & 31.3  & 0.58  & 0.646 \\
    \hline
  \onlenlp & mr2en-Method2-c & 30.6 & 0.57 & 0.659\\
    \hline
\onlenlp  & mr2en-Method1-c  & 20.7 & 0.48 & 0.735\\
  \hline
 \ucf   & mr2en-UnigramSegmentation-b  & 7.7 &  0.24 & 0.833\\
  \hline
\iiitt & mr2en-IndicTrans-b &  5.1 & 0.22 & 1.002\\
  \hline
    \end{tabular}
\caption{Results of submitted systems at English$\leftrightarrow$Irish \& English$\leftrightarrow$Marathi in the ``-b'' and ``-c'' method}
\label{tab:results2}
\end{table*}

\newpage

\section{Conclusion}
\label{sec:Conclusion}
We have reported the findings of the LoResMT 2021 Shared Task on COVID and sign language translation for low-resource languages as part of the fourth LoResMT workshop. All submissions used neural machine translation except for the one from \athree. We conclude that in our shared tasks the use of transfer learning, domain adaptation, and back translation achieve optimal results when the data sets are domain specific as well as small-sized. Our findings show that uni-gram segmentation transfer learning methods provide comparatively low results for the following metrics: BLEU, CHRF and TER. The highest BLEU scores achieved are 36.0 for English-to-Irish, 34.6 for Irish-to-English, 24.2 for English-to-Marathi, and 31.3 for Marathi-to-English.

In future iterations of the LoResMT shared tasks, extended corpora of the three language pairs will be provided for training and evaluation. Human evaluation on system results will also be conducted. For sign language MT, the tasks will be fine-grained and evaluated separately. 

\section{Acknowledgements}
This publication has emanated from research in part supported by Cardamom-Comparative Deep Models of Language for Minority and Historical Languages (funded by the Irish Research Council under the Consolidator Laureate Award scheme (grant number IRCLA/2017/129)) and we are grateful to them for providing English$\leftrightarrow$Irish parallel and monolingual COVID-related texts. We would like to thank Panlingua Language Processing LLP and Potamu Research Ltd for providing English$\leftrightarrow$Marathi parallel and monolingual COVID data and Taiwanese Sign Language$\leftrightarrow$Traditional Chinese linguistic data, respectively.

\newpage

\bibliographystyle{apalike}
\bibliography{mtsummit2021}

\end{document}